\documentclass[10pt,twocolumn]{article}
\usepackage[margin=0.75in]{geometry} % Adjust margins
\usepackage{times}  % DO NOT CHANGE THIS
\usepackage{helvet}  % DO NOT CHANGE THIS
\usepackage{courier}  % DO NOT CHANGE THIS
\usepackage[hyphens]{url}  % DO NOT CHANGE THIS
\usepackage{graphicx} % DO NOT CHANGE THIS
\usepackage{natbib}  % DO NOT CHANGE THIS AND DO NOT ADD ANY OPTIONS TO IT
\usepackage{caption} % DO NOT CHANGE THIS AND DO NOT ADD ANY OPTIONS TO IT
\usepackage{algorithm}
\usepackage{algorithmic}

\usepackage{array}
\usepackage{colortbl}
\usepackage{booktabs}
\usepackage{amssymb}
\usepackage{multirow}
\usepackage{amsmath}
\usepackage{caption}
\usepackage{subcaption}

\usepackage{newfloat}
\usepackage{listings}
\DeclareCaptionStyle{ruled}{labelfont=normalfont,labelsep=colon,strut=off} % DO NOT CHANGE THIS
\floatstyle{ruled}
\newfloat{listing}{tb}{lst}{}
\floatname{listing}{Listing}
\title{\textbf{Multiple Stochastic Prompt Tuning for Few-shot Adaptation under Extreme Domain Shift}}
\author{
    Debarshi Brahma and Soma Biswas \\
    Indian Institute of Science, Bangalore, India \\
    \{debarshib, somabiswas\}@iisc.ac.in
}
\date{}

\begin{document}

\maketitle

\begin{abstract}
Foundation Vision-Language Models (VLMs) like CLIP exhibit strong generalization capabilities due to large-scale pretraining on diverse image-text pairs. However, their performance often degrades when applied to target datasets with significant distribution shifts in both visual appearance and class semantics. Recent few-shot learning approaches adapt CLIP to downstream tasks using limited labeled data via adapter or prompt tuning, but are not specifically designed to handle such extreme domain shifts. 
Conversely, some works addressing cross-domain few-shot learning consider such domain-shifted scenarios but operate in an episodic setting with only a few classes per episode—limiting their applicability to real-world deployment, where all classes must be handled simultaneously.
To address this gap, we propose a novel framework, \textit{MIST} (Multiple Stochastic Prompt Tuning), for efficiently adapting CLIP to datasets with extreme distribution shifts using only a few labeled examples, in scenarios involving all classes at once.
Specifically, we introduce multiple learnable prompts per class to effectively capture diverse modes in visual representations arising from distribution shifts.
To further enhance generalization, these prompts are modeled as learnable Gaussian distributions, enabling efficient exploration of the prompt parameter space and reducing overfitting caused by limited supervision.
Extensive experiments and comparisons with state-of-the-art methods demonstrate the effectiveness of the proposed framework.
\end{abstract}
% ***********************

\section{Introduction}

Foundation Vision-Language Models (VLMs) such as CLIP~\cite{clip} and ALIGN~\cite{align} have significantly advanced computer vision by enabling strong zero-shot generalization across tasks. Trained on large-scale image-text pairs, they learn robust representation spaces. However, such pretraining is costly and impractical to replicate for every new task, prompting growing interest in efficient adaptation using only a few labeled examples.

However, few-shot adaptation of such large-scale models is challenging due to the risk of overfitting and the potential loss of their original pretrained generalization capabilities. To address this, an emerging body of work has explored parameter-efficient fine-tuning techniques, such as prompt tuning~\cite{coop} and adapter tuning~\cite{clip_adapter}. These methods avoid full model fine-tuning by introducing a small set of trainable parameters while keeping the backbone frozen—either by modifying the input space (prompt tuning) or the output layers (adapter tuning). Despite these advances, such approaches primarily focus on standard benchmark datasets with natural images and class semantics (e.g., ImageNet, Caltech101), and largely overlook scenarios involving extreme domain shifts. In real-world applications, downstream datasets can exhibit significant shifts in both visual appearance and label semantics. For example, medical image datasets often feature domain-specific content and class names that do not align with natural image concepts and are typically unavailable for pretraining due to privacy concerns.
While recent works have applied CLIP to cross-domain few-shot learning (CDFSL)~\cite{semantic_guided}, they typically adopt a source-free meta-testing setup, adapting to episodes with a few sampled classes (e.g., 5-way) from the target domain. Performance is averaged over many such episodes. However, this approach is computationally expensive and misaligned with real-world settings, where all target classes are present simultaneously.

In this work, we propose a novel prompt learning framework, \textit{MIST} (Multiple Stochastic Prompt Tuning), for adapting CLIP to a more realistic setting, where target datasets exhibit significant domain and semantic shifts, and only a few labeled examples from all classes are available simultaneously.
We first observe that extreme distribution shifts can lead to fragmented visual representations, forming separate and inconsistent clusters in the embedding space (Fig.~\ref{fig:multiple_tsne}). 
Moreover, large number of classes with semantic shifts (different class labels) can cause multiple class features to cluster together, resulting in ambiguous decision boundaries.
To address these challenges, we introduce multiple learnable prompts per class, enabling better modeling of multi-modal feature distributions. Further, instead of directly optimizing prompt weights, we represent each prompt as a Gaussian distribution with learnable mean and variance, promoting diverse and well-separated representations while mitigating overfitting through efficient exploration of the prompt space.
The key contributions of this work are summarized below:
%To address this gap, we propose a novel prompt learning framework, \textit{MIST (\underline{M}ult\underline{I}ple \underline{ST}ochastic Prompt Tuning)}, for adapting CLIP to a more realistic setting—where the target datasets exhibit significant domain and semantic shifts, along with few training examples from all classes simultaneously, as opposed to the constrained episodic structure. First, We observe that existing prompt tuning approaches, which operate under similar non-episodic protocols, often struggle to handle such extreme distribution shifts, since it can lead to visual features forming separate, inconsistent clusters in the representation space (Fig.~\ref{fig:multiple_tsne}). Additionally, the shift in label semantics can result in the class-specific features clustering  more closely in the representation space when all classes are present during adaptation, leading to harder-to-discriminate decision boundaries. Inspired by these observations, we introduce multiple learnable prompts per class to better model the multiple peaks in the visual representation space.
%Additionally, unlike standard optimization of prompt weights, we model the different prompts with Gaussian distributions represented by learnable mean and variance vectors. This approach encourages uniform separation of the class-specific features and mitigates overfitting, by facilitating efficient exploration of the prompt parameter space.
%The key contributions of this work can be summarized as follows:

\begin{enumerate}
\item We propose a novel framework for few-shot adaptation of CLIP to realistic scenarios involving extreme domain and label semantic shifts, with all target classes present simultaneously.
\item We analyze limitations of existing prompt tuning methods for few-shot setting, under severe distribution shifts.
\item We propose \textit{MIST}, a novel prompt tuning framework that uses multiple class-specific prompts to model multimodal visual feature distributions.
\item We further represent each prompt as a learnable Gaussian distribution, enabling better generalization and reducing overfitting in low-data regimes.
\item Extensive experiments demonstrate that MIST outperforms state-of-the-art methods across multiple challenging benchmarks.
\end{enumerate}
\label{sec:intro}

% ***************
\section{Related Work}
Here, we briefly discuss the related work in literature. \\
\textbf{Vision-language foundation models.} Recently, foundation Vision-Language Models (VLMs)~\cite{clip,align,blip} have shown strong zero-shot generalization by learning aligned visual-textual representations from web-scale image-text pairs. However, their performance can degrade on domain-specific tasks with distribution shifts or rare class semantics not seen during pretraining. As retraining such large models for new tasks is impractical, recent efforts focus on efficient adaptation using limited labeled samples from the target domain. \\
% The recent emergence of foundation Vision-Language Models (VLMs)~\cite{clip,align,blip} has significantly transformed the computer vision landscape, owing to their impressive zero-shot generalization capabilities. These models are trained on large web-scale image-text pairs using a contrastive learning framework, enabling them to learn rich, aligned visual and textual representations without the need for task-specific supervision. While these VLMs have shown strong performance across a wide range of downstream tasks, they may fall short in more specialized scenarios—particularly in domains where data distributions differ significantly from those seen during pretraining. Moreover, many real-world datasets contain domain-specific content and label semantics (e.g., medical or industrial imagery) that are unlikely to be well-represented in web-scale corpora. Since retraining or finetuning such large models for every new task is computationally prohibitive, recent research has focused on developing data--efficient adaptation techniques that can leverage only a few labeled samples from the target domain. \\
\textbf{Few-shot adaptation of VLMs.}
Adapting these large-scale models to downstream tasks with few labeled training data is often challenging, due to the risk of overfitting. Efficient transfer learning methods like prompt tuning~\cite{coop,maple,promptsrc} or adapter tuning~\cite{clip_adapter,tip_adapter}
address this issue by optimizing only a few parameters added to these models, either in the input space or in the intermediate or output layers. For instance, CLIP-Adapter~\cite{clip_adapter} trains a classifier on the visual output features to modify the visual feature space, and Tip-Adapter~\cite{tip_adapter} stores the few-shot image prototypes, which are used to compute similarity with the test samples to guide the visual encoder. In prompt-tuning, CoOp~\cite{coop} trains a few prompt vectors appended to the classname text, keeping the CLIP encoders frozen. MaPLe~\cite{maple} proposes training prompts in both the textual and visual branches to improve multimodal alignment, while PromptSRC~\cite{promptsrc} further enhances performance by distilling knowledge from the frozen CLIP model. \cite{tcp} incorporates class description features into the text encoder during prompt tuning to improve discriminability of the classifier. \cite{proda} tunes text prompts using a pool of diverse prompts for each class, while \cite{bayesian_pl} adds learnable gaussian noise to each token of the learnable text prompt, to improve generalization. These works mainly focus on standard image datasets with natural images and class semantics, often overlooking more realistic deployment scenarios where the target datasets may exhibit substantial domain shifts or unfamiliar, specialized label semantics. Some recent works have employed CLIP for the cross-domain few-shot learning (CDFSL) setup, which incorporates these challenges~\cite{semantic_guided, prompt_free}. However, these methods take the episodic paradigm, which samples fixed number of classes in each episode, which is often unrealistic in real-world deployment. \\
% \textit{Our method utilizes a multiple prompt learning framework, whose weights are sampled from different learnable Gaussian distributions to address the pCDFSL.} \\
%\textit{Our method utilizes a multimodal prompt learning framework similar to~\cite{maple,promptsrc}. Contrary to~\cite{proda,bayesian_pl}, we learn multiple prompt classifiers whose weights are sampled in two distinctive ways from different learnable Gaussian distributions.} \\
% ***********************
\textbf{Stochastic neural networks.} Standard neural networks train weights deterministically as point-estimates. Contrarily, Bayesian Neural Networks~\cite{bnns,weight_uncertainty} model the weights as probability distributions, making them useful in handling uncertainty in predictions as well as learning robust representations. Stochastic classifiers have been explored in UDA~\cite{star}, person re-identification~\cite{re-id}, incremental learning~\cite{s3c} and DG~\cite{stylematch} in prior literature. 
\textit{To the best of our knowledge, this is the first work which explores stochastic classifiers for few-shot adaptation of VLMs under extreme domain shifts.}

\section{Problem Formulation}
% \subsection{Practical CDFSL (pCDFSL) paradigm}
Given few labeled training examples from a target dataset, the task is to adapt the CLIP model efficiently to this data. 
% Cross-domain few-shot learning (CDFSL) based methods typically focus on classification in an episodic manner.
Formally, we have a support set $\mathcal{S} = \{(X_i, y_i)\}^{C\times k}_{i=1}$ from the target dataset containing $k$ samples from all the $C$ classes simultaneously. Here, $y_i\in\{0,1\}^C$ is the corresponding ground truth label, and $k=\{1,2,4,8,16\}$, denotes the number of shots. The evaluation is done on the full test set. Here, in addition to the few-shot problem, the target dataset contains significant domain and label semantic shift from natural image datasets.
\subsection{Preliminaries}
Here, we briefly describe the process of classifying images with CLIP and the base network used in this work for completion.
Let us denote the CLIP text and image encoders as $\mathcal{F}_t$ and $\mathcal{F}_v$ respectively. The input image $Xv = \mathbb R^{C\times H\times W}$ is broken up into patches $\{e_{CLS}, e_1, e_2, ..., e_M\}$ and fed into the image encoder to extract the image embedding $z_v = \mathcal{F}_v(X_v)$. Similarly, the text input (typically of the form ``A photo of a [CLS]") is tokenized in the form $X_t = \{t_{SOS}, t_1, t_2, ..., t_{CLS}, t_{EOS}\}$ and fed into the text encoder to get the text embedding $z_t = \mathcal{F}_t(X_t)$. During zero shot classification, the class text embeddings are matched with the image as follows: $\frac{exp(<z_v, z_t>/\tau)}{\sum\limits_{i=1}^{C}exp(<z_v, z_{t_i}>/\tau)}$, where $C$ is the number of classes and $\tau$ is the temperature constant. The class with highest similarity is the prediction. \\
% % ****************************
% \begin{figure}[t!]
%     \centering
%     \includegraphics[width=\linewidth]{ICCV2025-Author-Kit-Feb/figures/stochastic_prompt_learning1.pdf}
%     \caption{Effect of deterministic vs stochastic prompt learning on the EuroSAT dataset with 1-shot per class. The projected class-specific image features are spread in fixed regions for stochastic learning, implicitly learning well-separated decision boundaries.}
%     \label{fig:stochastic}
% \end{figure}

% %*******************************
\noindent
\textbf{Base Network of MIST:} 
In MIST, we employ a multimodal prompt learning strategy, where learnable prompt vectors are appended to the image and textual input branches. 
Specifically, let the learnable text prompt vectors be denoted as $\theta_t=\{\theta_{t_1},\theta_{t_2},...,\theta_{t_m}\}$ and the visual prompts as $\theta_v=\{\theta_{v_1},\theta_{v_2},...,\theta_{v_m}\}$. 
These are appended to the input text and image patches to form the modified inputs as: $\tilde X_t = \{t_{SOS}, \theta_{t_1},...,\theta_{t_m}, t_1, t_2, ..., t_{CLS}, t_{EOS}\}$ and $\tilde X_v = \{e_{CLS}, \theta_{v_1},...,\theta_{v_m}, e_1, e_2, ..., e_{M}\}$ respectively. The extracted feature embeddings from the CLIP encoders are now $\tilde z_t = \mathcal{F}_t(\tilde X_t)$ and $\tilde z_v = \mathcal{F}_v(\tilde X_v)$. Here, the trainable textual prompts are passed through a learnable projection layer $f_\phi$ to obtain the visual prompts, i.e., $\theta_v = f_\phi(\theta_t)$. Along with adding prompts to the inputs, we also adopt a deep prompting approach~\cite{maple,promptsrc}, where learnable prompt vectors are attached after every transformer block. When adapting to a downstream task, these multimodal prompts are trained in an end-to-end manner, keeping the CLIP encoders frozen.

\section{The Proposed Framework}

\begin{figure*}[t!]
    \centering
    \includegraphics[width=0.77\textwidth]{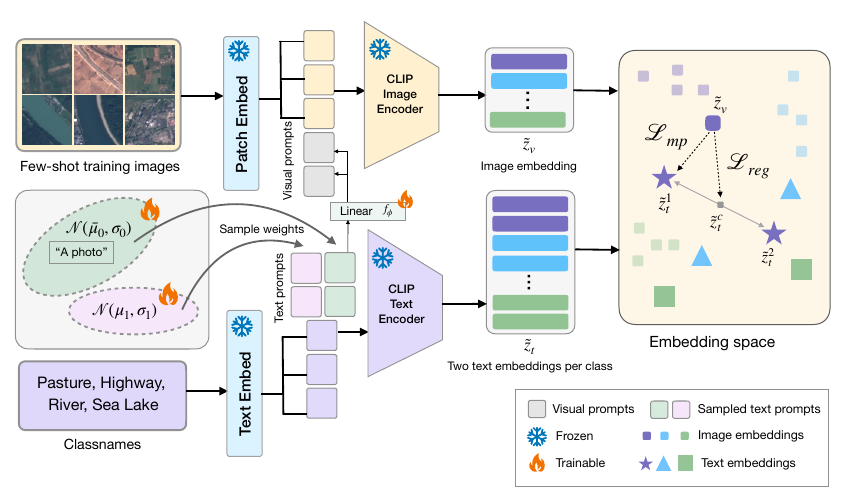}
    \caption{\textbf{Our proposed MIST framework.} We append two sets of prompts to the classnames, one sampled from a fixed mean ($\bar{\mu}_0,\sigma_0$) and the other from a fully learnable Gaussian distribution ($\mu_1,\sigma_1$). $f_\phi$ projects the text prompts to visual prompts. The loss term $\mathcal{L}_{mp}$ trains the distribution parameters ($\mu_1,\sigma_0,\sigma_1$) and $f_\phi$ such that the image embedding is assigned to the closest text prototype of its respective class. The $\mathcal{L}_{reg}$ term prevents the two prompts from collapsing by enforcing diversity in the class-wise prompt training.}
    \label{fig:main_figure}
\end{figure*}

The goal is to efficiently adapt the large-scale pre-trained CLIP using very few samples from all the classes simultaneously, under extreme domain and semantic shifts. 
To this end, we present MIST (Fig.~\ref{fig:main_figure}), which augments the base network with two novel modules to tackle these challenges:
(i) Stochastic prompt learning to mitigate the risk of overfitting due to very few training examples and (ii) Multiple prompts to address the implicit unimodal assumption regarding the class data distributions, which we descrie below. %Together they constitute our final proposed framework, which is illustrated in Figure~\ref{fig:main_figure}.
\begin{figure}[t!]
    \centering
    \includegraphics[width=\linewidth]{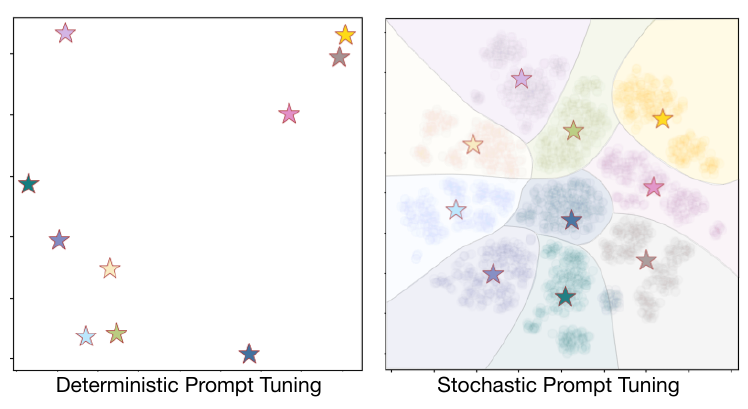}
    \caption{Effect of deterministic vs stochastic prompt learning on the EuroSAT dataset with 1-shot per class. The projected class-specific image features are spread in fixed regions for stochastic learning, implicitly learning well-separated decision boundaries.}
    \label{fig:stochastic}
\end{figure}

%*******************************

\begin{table}[t]
    \centering
    \small
    \begin{tabular}{lccccc}
    \toprule
    Method & EuroSAT & ISIC \\ 
    \midrule
    \multicolumn{3}{c}{1-Shot} \\ \midrule
    Deterministic Prompt Tuning & 73.30 & 27.50 \\ 
    Stochastic Prompt Tuning ($\mu, \sigma^*$) & 73.43 & 30.67 \\
    Stochastic Prompt Tuning ($\mu^*, \sigma^*$) & 67.40 & 22.67  \\
    \midrule
    \multicolumn{3}{c}{8-Shots} \\ \midrule
    Deterministic Prompt Tuning & 86.80 & 46.53 \\
    Stochastic Prompt Tuning ($\mu, \sigma^*$) & 86.73 & 50.80 \\
    Stochastic Prompt Tuning ($\mu^*, \sigma^*$) & 88.03 & 50.93 \\ 
    \bottomrule
    \end{tabular}
    \caption{Stochastic prompt learning with two different sampling techniques on EuroSAT and ISIC datasets. The fixed mean approach ($\mu,\sigma^*$) performs better in low shots, while sampling from a fully learnable distribution ($\mu^*,\sigma^*$) performs better in higher shots.}
    \label{tab:stochastic prompts}
\end{table}

% ****************************
\begin{figure}[t!]
    \centering
    \includegraphics[width=\linewidth]{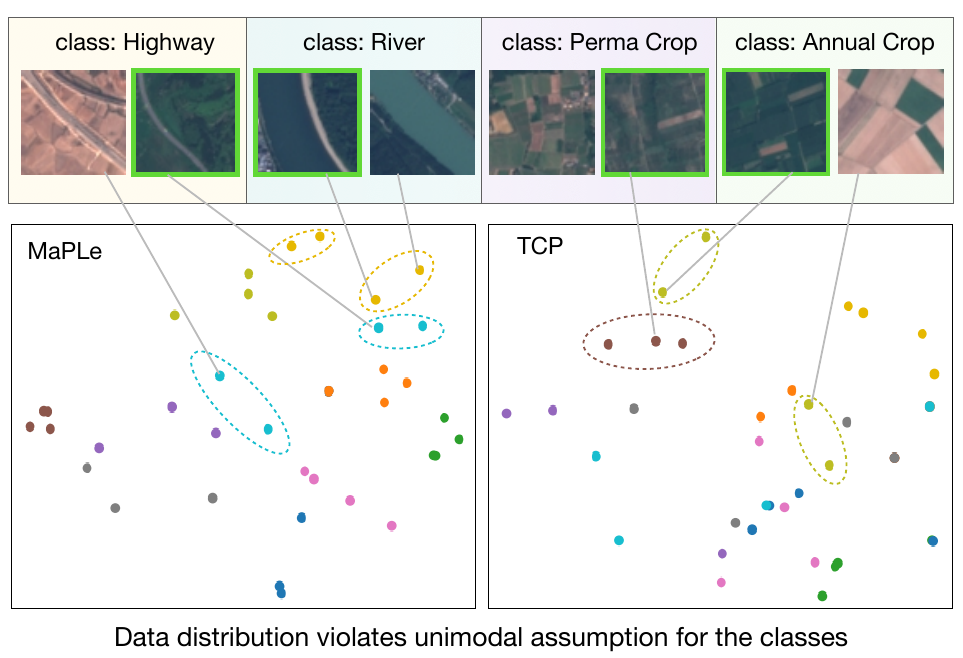}
    \caption{t-SNE visualization of image embeddings from MaPLe~\cite{maple} (left) and TCP~\cite{tcp} (right) for the EuroSAT dataset. The classwise data distribution violates the unimodal assumption, due to high interclass similarity and intraclass variations.}
    %, along with extreme domain shifts.}
    \label{fig:multiple_tsne}
\end{figure}

% ****************************
% ****************************
\subsection{Stochastic Prompt Learning}
\label{stochastic prompt learning}

Although existing prompt tuning methods optimize a few learnable parameters appended to the inputs,
% to increase the cosine similarity with the respective class image prototypes. 
%where we represent each class with two prompts, each being modeled as distinct Gaussian distributions with learnable mean and variance vectors. 
the limited availability of training samples makes large-scale models like CLIP prone to overfitting as observed by~\cite{promptsrc}. Additionally, when all target classes are present during adaptation, domain-specific or unfamiliar classnames can cause the class-specific features to lie closer together in the representation space, making inter-class boundaries harder to distinguish. To address these issues, we explore a novel strategy using stochastic prompt learning, where we model the prompts with learnable distributions.
% While prior methods have addressed this using regularization modules~\cite{promptsrc} or architectural modifications~\cite{maple}, here we explore a different strategy using stochastic prompt learning to address the overfitting issue. 

The main idea is to learn a distribution over the prompt parameters instead of optimizing them as point-estimates as is the standard practice. This helps to mitigate the uncertainty arising from scarce data, since each distinct sampled weight from this learnable distribution forms diverse decision boundaries for the few shot data, by allowing a richer exploration of the prompt parameter space. This provides an implicit regularization to the model without additional loss functions leading to more robust decision boundaries for the few-shot training samples.
Specifically, we sample the text prompt weights from a Gaussian distribution $\mathcal{N}(\mu,\sigma)$, parameterized by mean $\mu$ and variance $\sigma$ as follows: $\theta_t \sim \mathcal{N}(\mu,\sigma)$. After every iteration, we backpropagate the loss to the learnable $\mu$ and $\sigma$ parameters. For end-to-end training, we use the Gaussian reparameterization trick~\cite{var_auto_encoder} as follows:
\begin{equation}
    \theta_t = \mu + \mathcal{N}(0,I) \odot \sigma
\end{equation}
%We illustrate the effect of stochastic prompt learning in Fig.~\ref{fig:stochastic}. 
We observe from Fig.~\ref{fig:stochastic} that in the deterministic case, passing the same examples through the trained model always projects them to the same points in the feature space. In contrast, the projections in the stochastic scenario spreads over a broader region, due to sampling of weights from the learned distribution. 
This variability implicitly encourages a margin between the class-specific features, resulting in more discriminative decision boundaries.

To verify this, we consider two distinct strategies for sampling the text prompt parameters. First, we fix the mean of the Gaussian to the standard prompt ``A photo", keeping the variance learnable. In the second case, both the mean and variance are learnable. The results on two datasets~\cite{bscdfsl} are shown in Table~\ref{tab:stochastic prompts}.
We observe that in low-shot setting, the first approach outperforms the second, while an opposite trend is seen in the higher-shot setting. This suggests that with very few training data (1-shot), directly optimizing the parameters of a distribution is challenging, but exploring variations around a standard prompt improves performance. 
On the contrary, learning the full distribution with more data outperforms the fixed mean approach. Hence, the two strategies complement each other in facilitating an efficient coverage of the prompt parameter space.
% We also illustrate the effect of stochastic prompt learning in Figure~\ref{fig:stochastic}. Here, we pass the same samples through the trained model and plot the 
% \begin{figure}[t!]
%     \centering
%     \includegraphics[width=\linewidth]{ICCV2025-Author-Kit-Feb/figures/multiple_tsne_eurosat.pdf}
%     \caption{t-SNE visualization of image embeddings from MaPLe~\cite{maple} (left) and TCP~\cite{tcp} (right) for the EuroSAT dataset. The classwise data distribution violates the unimodal assumption, due to high interclass similarity and intraclass variations.}
%     %, along with extreme domain shifts.}
%     \label{fig:multiple_tsne}
% \end{figure}
% see that this approach learns more robust decision boundaries, compared to the deterministic approach where image features are projected onto decision boundaries, highlighting uncertain decision regions.
% \color{red} A real/toy example to illustrate the point will be good\color{black}

% **************************
\subsection{MIST: Multiple Stochastic Prompt Tuning}
\label{msp}
Since CLIP is pretrained on web-scale data encompassing mainly natural images~\cite{clip}, the image encoder struggles to learn robust classwise features when faced with extreme domain and semantic shifts in the target dataset. In addition, intraclass diversity and interclass similarity in the images due to presence of many classes, results in disjoint clusters in visual features from the same class. 
Consider an illustrative example of the EuroSAT dataset containing satellite images of various terrains. Here, distinct classes like ``Highway" and ``River" can have similar visual representatives in the few-shot training data, at the same time featuring diverse visuals from the same classes as shown in Fig.~\ref{fig:multiple_tsne}. Existing prompt tuning approaches represent each class with single learnable prompts, implicitly assuming that classwise visual features form single clusters. However, such a strategy is insufficient to represent the disjoint visual clusters, which may result from such challenging settings. To illustrate this, we consider two representative prompt-tuning methods, MaPLe~\cite{maple} and TCP~\cite{tcp} and show their t-SNE visualizations in Fig.~\ref{fig:multiple_tsne}. We observe that the unimodal assumption is violated and the image embeddings of each class form multiple modes in the representation space.

To address this, inspired from~\cite{imp,mixtfsl}, we introduce a multiple prompt learning approach, where we represent each class with multiple prompt vectors. However, incorporating too many learnable prompts for each class is not desirable since:
(i) Each class contains few training samples, and many text embeddings per class could lead to overfitting on individual datapoints, resulting in loss of class level representations, and (ii) it introduces additional learnable parameters and thus more computational overhead. As a balanced approach, we represent each class with two prompt vectors.
Formally, let the embeddings corresponding to the two prompts for a particular class $CLS_k$ be denoted as $\tilde z_t^i$, where, $i$ = 1, 2. Here, $\tilde z_t^i = \mathcal{F}_t(\tilde X_t^i)$, where, $\tilde X_t^i = \{t_{SOS},\theta_t^i,t_1,...,t_{CLS_k},t_{EOS}\}$ is the $i^{th}$ text prompt for the class $CLS_k$. $\theta_t^i=\{\theta_{t1}^i, \theta_{t2}^i,...,\theta_{tm}^i\}$ denotes the $i^{th}$ set of learnable prompt vectors for that particular class.

To simultaneously represent the underlying multimodal class distribution and mitigate overfitting, we stochastically model the parameters of the two prompts for each class as described in Sec~\ref{stochastic prompt learning}. Specifically, we incorporate the fixed mean, learnable variance approach on the parameters of the first prompt, simultaneously learning a full Gaussian distribution over the parameters of the second prompt:
\begin{equation}
    \begin{aligned}
        \theta_t^1 \sim \mathcal{N}(\bar{\mu}_0,\sigma_0) \\
        \theta_t^2 \sim \mathcal{N}(\mu_1,\sigma_1)
    \end{aligned}
\end{equation}
Here, $\bar{\mu}_0$ is a fixed vector corresponding to the text embedding of ``A photo", and $\mu_1$, $\sigma_0$, $\sigma_1$ are learnable parameters. 
Now, we describe the training process of MIST. \\ %Sampling the two prompts from two learnable distributions has the risk of learning the same distributions, effectively collapsing the two prompts into a single prompt \color{red} []\color{black}. The proposed approach ensures a distinctive learning framework, facilitating a richer coverage of the prompt parameter space as we observed in Sec~\ref{stochastic prompt learning}. \\
\noindent
\textbf{MIST Training:} For a particular image embedding $\tilde z_v$, we first find the closest text embedding of its respective class after every iteration, based on cosine similarity as follows:
\begin{equation}
    i^* = \operatorname*{argmax}_{i \in \{1,2\}} \,\,\, sim(\tilde z_t^i, \tilde z_v)
\end{equation}
where, $sim(a,b)=\frac{a\cdot b}{\lVert a\rVert\lVert b\rVert}$ denotes the cosine similarity. 
During training, the image embedding is assigned to its closest prompt embedding by minimizing the following loss function:
%The training is done based on the above equation to assign $\tilde z_v$ to its closest prompt embedding by minimizing the following loss function:
% \begin{equation}
%     \mathcal{L}_{mp} = -\frac{1}{nC}\sum\limits_{j=1}^{nC}y_j\, log(sim(\tilde z_t^j, \tilde z_v))
% \end{equation}
\begin{equation}
    \mathcal{L}_{mp} = - log\,\Big(\frac{exp(sim(\tilde z_t^{i^*}, \tilde z_v))}{\sum\limits_{j=1}^{2C}exp(sim(\tilde z_t^j, \tilde z_v))}\Big)
\end{equation}
% where, $y_j$ is the one-hot encoding with $1$ at $j=i^*$, and $sim(\cdot)$ denotes the cosine similarity.
where, $C$ is the number of classes, and $sim(\cdot)$ denotes the cosine similarity.
To prevent the image embedding $\tilde z_v$ from being always assigned to a single text prompt, and encourage diversity when training the two prompts within the same class, we minimize an additional regularization term to increase the cosine similarity of the image embedding to the centroid of the two text embeddings of its corresponding class:
\begin{equation}
    \mathcal{L}_{reg} = - sim(\tilde z_v, \tilde z_t^c)
\end{equation}
where, $\tilde z_t^c=\frac{1}{2}(\tilde z_t^1+\tilde z_t^2)$ is the centroid of the prompt embeddings for the corresponding class and $sim(\cdot)$ represents the cosine similarity. Thus, the final objective function is:
\begin{equation}
    \mathcal{L}_{total} = \mathcal{L}_{mp} + \mathcal{L}_{reg}
\end{equation}
This loss function is used to optimize the Gaussian parameters $\mu_1$, $\sigma_0$ and $\sigma_1$, as well as the projection layers as:
\begin{equation}
    \mu_1^*,\sigma_0^*,\sigma_1^*,\phi^* = \operatorname*{argmin}_{\mu_1,\sigma_0,\sigma_1,\phi} \mathbb E_{(X,y)\sim D_{tgt}}  \mathcal{L}_{total}(X,y)
\end{equation}
\begin{table}[t]
\centering
\footnotesize
\setlength{\tabcolsep}{4pt}
\scalebox{0.77}{
\begin{tabular}{lccccc}
\toprule
   Method      & EuroSAT & ISIC  & PDisease & ChestX & Average \\ \midrule
   & \multicolumn{5}{c}{1-shot} \\ \midrule
         
  CoOp \scriptsize (IJCV'22)    & $51.87$   & $22.77$ & $24.73$        & $\textbf{22.83}$  & $30.55$   \\
        TaskRes \scriptsize (CVPR'23) & $64.67$   & $19.70$  & $36.57$    & $10.97$  & $32.98$   \\
         MaPLe \scriptsize (CVPR'23) & $73.30$    & $27.50$  & $51.53$        & $14.60$   & $41.73$ \\
       PromptSRC \scriptsize (ICCV'23)  & $73.23$   & $21.97$ & $\textbf{55.03}$        & $14.37$  & $41.15$   \\
        CLAP \scriptsize (CVPR'24)  & $61.46$   & $26.61$ & $47.22$   & $15.94$  & $37.81$   \\
        TCP \scriptsize (CVPR'24)  & $64.30$    & $27.80$    & $49.37$      & $14.93$  & $39.10$ \\
       MIST (Ours)    & $\textbf{77.90}$    & $\textbf{34.40}$  & $50.27$     & $17.10$   & $\textbf{44.92}$   \\ \midrule
        & \multicolumn{5}{c}{2-shot} \\ \midrule
CoOp \scriptsize (IJCV'22)   & $66.00$      & $21.87$ & $37.97$    & $14.43$  & $35.07$   \\
       TaskRes \scriptsize (CVPR'23) & $68.83$   & $23.13$ & $39.27$    & $10.83$  & $35.52$   \\
         MaPLe \scriptsize (CVPR'23)  & $78.07$   & $31.90$  & $67.17$ & $16.27$  & $48.35$  \\
         PromptSRC \scriptsize (ICCV'23)   & $79.53$   & $29.47$ & $68.07$    & $12.70$   & $47.44$   \\
        CLAP \scriptsize (CVPR'24)   & $70.63$   & $34.79$ & $60.13$    & $\textbf{16.43}$  & $45.50$   \\
        TCP  \scriptsize (CVPR'24)  & $70.37$   & $\textbf{36.87}$ & $62.63$     & $15.63$   & $46.38$   \\
       MIST (Ours)    & $\textbf{81.57}$   & $36.37$ & $\textbf{69.60}$   & $13.90$   & $\textbf{50.36}$   \\ \midrule
        & \multicolumn{5}{c}{4-shot} \\ \midrule
CoOp \scriptsize (IJCV'22)   & $66.53$   & $25.00$   & $42.67$ & $17.93$  & $38.03$   \\
      TaskRes \scriptsize (CVPR'23) & $72.40$    & $21.40$  & $39.35$     & $10.27$  & $35.86$   \\
       MaPLe \scriptsize (CVPR'23)  & $84.03$   & $37.17$ & $77.07$      & $\textbf{19.73}$  & $54.50$    \\
        PromptSRC \scriptsize (ICCV'23)   & $85.23$   & $37.63$ & $78.70$     & $15.17$  & $54.18$   \\
         CLAP \scriptsize (CVPR'24)   & $76.43$   & $34.37$ & $65.11$   & $18.98$  & $48.72$   \\
      TCP \scriptsize (CVPR'24)    & $76.77$   & $37.37$ & $67.97$    & $17.07$  & $49.80$   \\
       MIST (Ours)    & $\textbf{85.93}$   & $\textbf{40.90}$  & $\textbf{79.67}$  & $18.67$  & $\textbf{56.29}$   \\ \midrule
        & \multicolumn{5}{c}{8-shot} \\ \midrule
CoOp \scriptsize (IJCV'22)   & $76.53$   & $38.27$ & $60.50$   & $14.60$   & $47.48$   \\
        TaskRes \scriptsize (CVPR'23) & $74.63$   & $34.30$  & $57.77$  & $12.57$  & $44.82$   \\
      MaPLe \scriptsize (CVPR'23)  & $86.80$    & $46.53$ & $84.47$   & $14.17$  & $57.99$   \\
    PromptSRC \scriptsize (ICCV'23)  & $88.37$   & $42.47$ & $86.80$   & $14.93$  & $58.14$   \\
       CLAP \scriptsize (CVPR'24)    & $76.85$   & $42.81$ & $74.62$   & $14.97$  & $52.31$   \\
     TCP \scriptsize (CVPR'24)    & $79.03$    & $46.57$  & $76.33$     & $14.97$  & $54.23$   \\
    MIST (Ours)    & $\textbf{88.63}$   & $\textbf{52.70}$  & $\textbf{87.47}$   & $\textbf{16.50}$   & $\textbf{61.33}$   \\ \midrule
        & \multicolumn{5}{c}{16-shot} \\ \midrule
CoOp \scriptsize (IJCV'22)   & $82.83$   & $43.40$  & $69.90$  & $\textbf{18.80}$   & $53.73$   \\ 
     TaskRes \scriptsize (CVPR'23) & $79.90$    & $38.10$  & $69.40$  & $12.87$  & $50.07$   \\
       MaPLe \scriptsize (CVPR'23)  & $92.80$    & $55.53$ & $89.93$  & $13.90$   & $63.04$   \\
       PromptSRC \scriptsize (ICCV'23)   & $92.55$   & $55.17$ & $91.40$  & $14.83$  & $63.49$   \\
       CLAP \scriptsize (CVPR'24)   & $82.96$ & $49.43$ & $78.27$  & $17.47$  & $57.03$   \\
       TCP \scriptsize (CVPR'24)    & $84.93$    & $52.83$  & $80.63$ & $16.53$  & $58.73$   \\
      MIST (Ours)    & $\textbf{93.57}$   & $\textbf{60.30}$  & $\textbf{91.73}$ & $14.77$  & $\textbf{65.09}$   \\
         \bottomrule
\end{tabular}
}
\caption{Performance comparison (average accuracy (\%) over 3 seeds) of the proposed MIST with the state-of-the-art approaches for $k=1,2,4,8,16$ shots from each class.}
\label{tab:main_results}
\end{table}
%****************************
\noindent
\textbf{Inference:} After learning the parameters of the distribution, during inference, we can sample weights for the two text prompts as follows: $\theta_t^1\sim \mathcal{N}(\bar{\mu}_0,\sigma_0^*)$ and $\theta_t^2\sim \mathcal{N}(\mu_1^*,\sigma_1^*)$. For each class, we take the maximum logit among the two text prompts as the output prediction for that class.

\section{Experimental Results}
Here, we extensively evaluate the proposed framework and compare it with the state-of-the-art approaches.
\noindent
% %****************************
\textbf{Datasets used:} For the target datasets, we consider the BSCDFSL~\cite{bscdfsl} benchmark, which is collected from real-world settings, and consists of four datasets, namely EuroSAT~\cite{eurosat}, ISIC~\cite{isic}, Plant Disease~\cite{plant_disease} and ChestX~\cite{chestx}. These datasets cover a varying spectrum of domain shifts, along with specialized classnames, encompassing satellite, agricultural and medical images. For training, we consider few samples ($k=1,2,4,8,16$) randomly selected from all the classes together and then evaluate the trained model on the full test set of all the datasets. The final accuracy is taken as the average over 3 different seeds. \\
% *************************
\textbf{Implementation details:} We employ CLIP ViT-B/16 as the backbone similar to MaPLe~\cite{maple}. The learnable context length of the text and vision inputs are set as $2$, and deep prompts are incorporated upto a depth of $9$. The model is trained using SGD optimizer for $150$ epochs with a learning rate of $0.0035$ and a batch size of $4$. All experiments are conducted on a NVIDIA RTX A5000 GPU.
% *************************
\subsection{Comparison with state-of-the-art methods}
To validate the effectiveness of our approach, we compare our proposed MIST with several recent CLIP-based efficient transfer learning methods 
%like CoOp, TaskRes, MaPLe, PromptSRC, CLAP and TCP, 
for varying number of shots. Specifically, we compare with 1) {\bf CoOp}~\cite{coop} and \textbf{TCP}~\cite{tcp}, which employ prompt tuning on the text branch; 
2) {\bf MaPLe}~\cite{maple} and \textbf{PromptSRC}~\cite{promptsrc} which utilize a multimodal prompt tuning approach;
3) {\bf TaskRes}~\cite{taskres} where task-specific adapters are tuned keeping the base text classifier fixed; 
4) {\bf CLAP}~\cite{clap} uses a linear-probing approach and mainly addresses the absence of validation sets in FSL.

For fair comparison, we run all the methods (using the official, publicly available codes) on the ViT-B/16 backbone and report the results in Table~\ref{tab:main_results}. 
We list some observations below: \\
(i) Among the competing methods, MaPLe and PromptSRC achieves the highest performance on average, closely followed by TCP. Their improved performance suggests the effectiveness of multimodal prompt tuning in handling distribution shifts over text prompt tuning, which was also observed in~\cite{maple}. This observation is further supported by our results; \\(ii) Although, CLAP is a recent approach, it mainly focuses on the validation problem of FSL. The reduced performance of CLAP highlights the limitations of linear probing, which does not utilize the text information for handling significant semantic and domain shifts; \\
(iii) As the number of shots increases, the multimodal prompt tuning approaches like MaPLe, PromptSRC and MIST outperforms other methods by larger margins, suggesting that training more parameters is more effective for higher shots. \\
(iv) Although all the methods show similar performance on the ChestX dataset, their overall accuracies are extremely low due to the large domain shift. However, text prompt tuning methods perform slightly better than the multimodal counterparts. This maybe because ChestX contains greyscale images, where additional prompt tuning in the vision branch degrades the performance. 
Overall, our proposed MIST framework outperforms the other methods significantly, giving consistent average gains of  $3.19\%$, $2.01\%$, $1.79\%$, $3.19\%$, $1.60\%$ on $k=1,2,4,8,16$ shots respectively, over the best performing methods. The significant improvement for the 1-shot case highlights the effectiveness of our approach in mitigating overfitting in extremely low data scenarios.
% %****************************
% \begin{table}[t]
%     \centering
%     \begin{tabular}{lcc}
%     \toprule
%         & EuroSAT & ISIC \\ \midrule
%         Maple & $72.90$ & $14.80$ \\ 
%         PromptSRC & $74.10$ & $16.10$ \\ 
%         TCP & $57.80$ & $13.13$ \\ 
%         MIST (Ours) & $\textbf{76.90}$ & $\textbf{16.73}$ \\ \bottomrule
%     \end{tabular}
%     \caption{Class imbalance in few-shot setting.}
%     \label{class_imbalance}
% \end{table}
% %*********************
%****************************
\begin{table}[t]
    \centering
    \scalebox{0.9}{
    \begin{tabular}{lcccc}
    \toprule
        Dataset & MaPLe & PromptSRC & TCP & MIST (Ours) \\ \midrule
        EuroSAT & $72.90$ & $74.10$ & $57.80$ & $\textbf{76.90}$ \\ 
        ISIC & $14.80$ & $16.10$ & $13.13$ & $\textbf{16.73}$ \\ \bottomrule
    \end{tabular}
    }
    \caption{Performance comparison (\%) of MIST with state-of-the-art methods on the class-imbalanced setting, with varying data samples from each class.}
    \label{class_imbalance}
\end{table}
%*********************
% ****************************
\begin{figure}[t]
    \centering
    \includegraphics[width=0.9\linewidth]{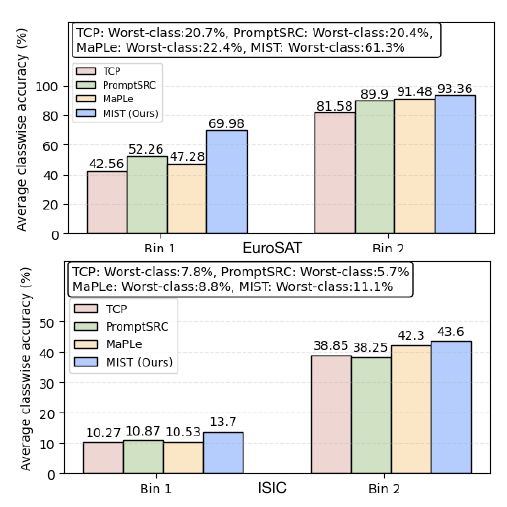}
    \caption{Generalization to classes: The class-wise accuracies are sorted and divided into 2 bins. Our proposed MIST outperforms the other methods in both the bins and also increases the worst-class accuracy for the same seed in the challenging 1-shot setting.}
    \label{fig:class_performance}
\end{figure}
% ****************************
\subsection{Additional Analysis}
Here we perform additional analysis and ablation studies to further validate our proposed framework. 
For the analysis, we compare with MaPLe, PromptSRC since they use multimodal prompts and are better suited for this task and TCP, since it is the state-of-the-art prompt tuning approach on CLIP. \\
%*********************
\textbf{1) Class-imbalanced learning:} Here, we explore an even more challenging scenario, where the number of labeled examples may vary across classes, reflecting real-world datasets. The standard few-shot settings in literature assume an idealistic scenario where each class has exactly $k$ training samples, overlooking the effect of class imbalance. To create such a setting, we perform data sampling in a cyclic manner, e.g., we take $\{1,2,4,8,1,2,...\}$ from each class of the target dataset for training. The model is then evaluated on the entire test set. The results on two representative datasets, EuroSAT and ISIC in Table~\ref{class_imbalance} shows that the proposed MIST outperforms the other methods even under class-imbalanced conditions, highlighting its effectiveness. \\ \\
% **********************
\textbf{2) Sensitivity to training samples:} The representation learning capability of a model is largely influenced by the {\em few} sampled training examples across classes. However, a robust model should ideally achieve a lower variance across different sampling strategies. To account for this, all the reported results are accuracies averaged over 3 different random seeds. Here, we further report the variance across the 3 seeds for the EuroSAT and ISIC datasets in Table~\ref{variance_seeds}. We observe that the proposed MIST not only achieves a higher accuracy, but also shows a much lower variance, highlighting its robustness to different sampling techniques. \\ \\
\noindent
\textbf{3) Generalization to all classes:} In practical scenarios, the overall accuracy is often not a reliable metric to understand the model's ability to represent difficult classes. Here, we study the effectiveness of our multiple stochastic prompt-tuning approach in learning generalized class boundaries and modeling all the complex class distributions.
%, using the worst-class accuracy in addition to the average class accuracy on the target datasets.
% **********************
\begin{table}[t]
    \centering
    \footnotesize
    \resizebox{\columnwidth}{!}{%
    \begin{tabular}{llccc}  % Fixed column format
        \toprule
        & &  EuroSAT &  ISIC \\ \midrule
        1-shot 
        &  MaPLe & \scriptsize $73.30$ \small $\pm 3.84$ & \scriptsize $27.50$ \small $\pm 10.19$  \\ 
        &  PromptSRC & \scriptsize $73.23$ \small $\pm 3.75$ & \scriptsize $21.97$ \small $\pm 6.09$  \\ 
        &  TCP & \scriptsize $64.30$ \small $\pm 3.24$ & \scriptsize $27.80$ \small $\pm 6.55$  \\ 
        &  MIST (Ours) & \scriptsize $77.90$ \small $\pm \textbf{2.63}$ & \scriptsize 34.40 \small $\pm \textbf{3.56}$ \\ \midrule
        2-shots &  MaPLe & \scriptsize $78.07$ \small $\pm 5.87$ & \scriptsize $31.90$ \small $\pm 6.08$ & \\ 
        &  PromptSRC & \scriptsize $79.53$ \small $\pm 2.76$ & \scriptsize $29.47$ \small $\pm 6.50$  \\ 
        &  TCP & \scriptsize $70.37$ \small $\pm 2.31$ & \scriptsize $36.87$ \small $\pm 8.29$ \\ 
        &  MIST (Ours) & \scriptsize $81.57$ \small $\pm \textbf{1.84}$ & \scriptsize $36.37$ \small $\pm \textbf{3.96}$ \\ \bottomrule
    \end{tabular}
    }
    \caption{MIST exhibits significantly lower variance across three different random seeds compared to other approaches.}
    \label{variance_seeds}
\end{table}
% **********************
Specifically, we first sort the class-wise accuracies in ascending order. 
The classes are then divided into two bins in this order to highlight the gain in accuracy in both the lower as well as the higher bin.
The comparison with the other methods are shown in Figure~\ref{fig:class_performance} for one random seed (same for all methods). We observe that the proposed MIST improves the accuracies in both the bins, while also increasing the worst-class accuracy, which indicates that MIST learns more generalized class representations.
\\ \\
\textbf{4) Ablation Study:}
Our proposed MIST framework models the two text classifiers using two distinct Gaussian sampling techniques as described earlier. 
Here we analyze the effectiveness of each of the proposed components in Table~\ref{ablation_table}. 
Our base method is the multimodal prompt tuning framework, MaPLe. Introducing stochasticity to this single prompt (fixed mean) tuning approach by sampling from a learnable Gaussian distribution mitigates overfitting as described earlier, and improves performance. 
%Here, we take the fixed mean approach due to 1-shot case. 
Introducing the second learnable prompt sampled from a fully learnable distribution without the $\mathcal{L}_{reg}$ term increases performance in EuroSAT, but reduces for ISIC. This can happen due to collapse of the two classifiers without the regularization term~\cite{mixtfsl, deepfake}. Finally, adding the regularization term $\mathcal{L}_{reg}$ outperforms the baseline  significantly for both the datasets. \\

\begin{figure}
    \centering
    \includegraphics[width=0.75\linewidth]{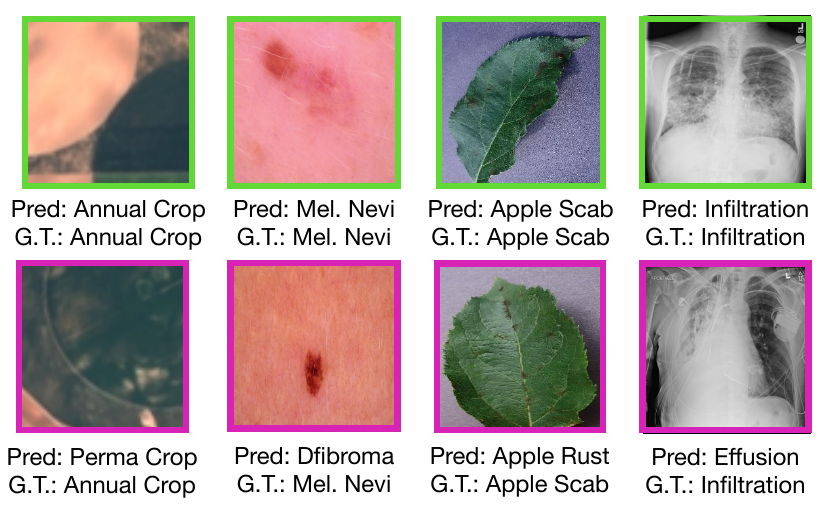}
    \caption{Qualitative results: From left to right, shows predictions on EuroSAT, ISIC, Plant Disease and ChestX respectively. Green denotes correct while red denotes incorrect predictions.}
    \label{fig:qual_results}
\end{figure}

% ****************************
\begin{table}[t]
    \centering
    \footnotesize
    \setlength{\tabcolsep}{3pt}
    \begin{tabular}{lcc}
    \toprule
        Method & EuroSAT & ISIC \\ \midrule
        
        Base Network & $73.30$ & $27.50$ \\ 
        + Stochastic Prompt Learning ($\mu$, $\sigma^*$) & $73.43$ & $30.67$ \\
        + Multiple Stochastic Prompting & $74.27$ & $27.00$ \\
        + Multiple Stochastic Prompting + $\mathcal{L}_{reg}$ (MIST) & $\textbf{77.90}$ & $\textbf{34.40}$ \\ \bottomrule
    \end{tabular}
    \caption{Ablation study (1-shot): All the proposed components collectively enhance the overall performance.}
    \label{ablation_table}
\end{table}
% ****************************
\begin{figure}[t]
    \centering
    \includegraphics[width=0.7\linewidth]{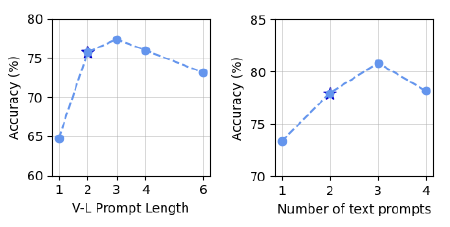}    
    \caption{Effect of prompt length (left) and number of text prompts per class (right) for EuroSAT data (1-shot).}
    \label{fig:ablations_combined}
\end{figure}
% ****************************
\noindent
\textbf{5) Number of text prompts \& Prompt Length:} MIST utilizes two prompts per class sampled from learnable Gaussian distributions. Here we study the effect of adding more text prompts. For this, we keep one prompt with a fixed mean, and the others sampled from fully learnable distributions.
From Figure~\ref{fig:ablations_combined} (right), we observe that the performance starts decreasing after three prompts, suggesting overfitting from the increasing number of learnable parameters. Further, addition of more classifiers introduces increased computational overhead and longer training time. 
% We have used 2 prompts for all the experiments.
The effect of increasing the number of learnable prompt vectors is illustrated in Figure~\ref{fig:ablations_combined} (left). 
Here also, the accuracy decreases after a point, indicating overfitting in the few-shot setting. We used 2 learnable prompts for all the experiments. \\ \\
% \color{red} We used \color{black}\\
\textbf{Qualitative Results:} We illustrate the inherent challenges of these datasets in Figure~\ref{fig:qual_results}, along with some of the predictions from the proposed MIST framework. \\
% ****************************

%\subsection{Future Directions}
\noindent
\textbf{Limitations.} While MIST outperforms state-of-the-art methods across all datasets, its performance slightly drops on the grayscale ChestX dataset, likely because of the additional visual prompts. In such cases, methods relying solely on textual prompts may prove more effective.
\section{Conclusion}
In this work, we propose a novel framework, MIST for adapting foundation VLMs like CLIP to realistic few-shot scenarios characterized by extreme domain and label semantic shifts. Motivated by the limitations of existing parameter efficient fine-tuning approaches, we incorporate multiple text prompts per class, modeled by distinct learnable Gaussian distributions to represent the inherent multimodal class distributions as well as mitigate overfitting. Extensive experiments on multiple benchmarks as well as additional analysis show the effectiveness of our proposed approach compared to state-of-the-art methods.

% propose a novel framework, MIST which incorporates multiple prompts per class, modeled by learnable Gaussian distributions to address the  
% Extensive experiments show the effectiveness of our proposed approach compared to state-of-the-art methods.
% In this work, we advocate a more realistic and challenging setting termed as the practical CDFSL (pCDFSL) setting, where a large-scale model like CLIP is adapted to a few-shot downstream dataset without the episodic constraints. We further propose a novel prompt tuning approach, MIST to address the various challenges of this setting. Different from existing prompt tuning approaches, our proposed MIST learns 

% \bibliography{aaai2026}

\bibliographystyle{plain} % or unsrt, or IEEEtran
\bibliography{references}

\begin{thebibliography}{10}

\bibitem{mixtfsl}
Arman Afrasiyabi, Jean-Fran{\c{c}}ois Lalonde, and Christian Gagn{\'e}.
\newblock Mixture-based feature space learning for few-shot image classification.
\newblock In {\em Proceedings of the IEEE/CVF international conference on computer vision}, pages 9041--9051, 2021.

\bibitem{imp}
Kelsey Allen, Evan Shelhamer, Hanul Shin, and Joshua Tenenbaum.
\newblock Infinite mixture prototypes for few-shot learning.
\newblock In {\em International conference on machine learning}, pages 232--241. PMLR, 2019.

\bibitem{weight_uncertainty}
Charles Blundell, Julien Cornebise, Koray Kavukcuoglu, and Daan Wierstra.
\newblock Weight uncertainty in neural network.
\newblock In {\em International conference on machine learning}, pages 1613--1622. PMLR, 2015.

\bibitem{isic}
Noel Codella, Veronica Rotemberg, Philipp Tschandl, M~Emre Celebi, Stephen Dusza, David Gutman, Brian Helba, Aadi Kalloo, Konstantinos Liopyris, Michael Marchetti, et~al.
\newblock Skin lesion analysis toward melanoma detection 2018: A challenge hosted by the international skin imaging collaboration (isic).
\newblock {\em arXiv preprint arXiv:1902.03368}, 2019.

\bibitem{bayesian_pl}
Mohammad~Mahdi Derakhshani, Enrique Sanchez, Adrian Bulat, Victor G~Turrisi da~Costa, Cees~GM Snoek, Georgios Tzimiropoulos, and Brais Martinez.
\newblock Bayesian prompt learning for image-language model generalization.
\newblock In {\em Proceedings of the IEEE/CVF International Conference on Computer Vision}, pages 15237--15246, 2023.

\bibitem{clip_adapter}
Peng Gao, Shijie Geng, Renrui Zhang, Teli Ma, Rongyao Fang, Yongfeng Zhang, Hongsheng Li, and Yu~Qiao.
\newblock Clip-adapter: Better vision-language models with feature adapters.
\newblock {\em International Journal of Computer Vision}, 132(2):581--595, 2024.

\bibitem{bscdfsl}
Yunhui Guo, Noel~C Codella, Leonid Karlinsky, James~V Codella, John~R Smith, Kate Saenko, Tajana Rosing, and Rogerio Feris.
\newblock A broader study of cross-domain few-shot learning.
\newblock In {\em Computer Vision--ECCV 2020: 16th European Conference, Glasgow, UK, August 23--28, 2020, Proceedings, Part XXVII 16}, pages 124--141. Springer, 2020.

\bibitem{eurosat}
Patrick Helber, Benjamin Bischke, Andreas Dengel, and Damian Borth.
\newblock Eurosat: A novel dataset and deep learning benchmark for land use and land cover classification.
\newblock {\em IEEE Journal of Selected Topics in Applied Earth Observations and Remote Sensing}, 12(7):2217--2226, 2019.

\bibitem{align}
Chao Jia, Yinfei Yang, Ye~Xia, Yi-Ting Chen, Zarana Parekh, Hieu Pham, Quoc Le, Yun-Hsuan Sung, Zhen Li, and Tom Duerig.
\newblock Scaling up visual and vision-language representation learning with noisy text supervision.
\newblock In {\em International conference on machine learning}, pages 4904--4916. PMLR, 2021.

\bibitem{s3c}
Jayateja Kalla and Soma Biswas.
\newblock S3c: Self-supervised stochastic classifiers for few-shot class-incremental learning.
\newblock In {\em European Conference on Computer Vision}, pages 432--448. Springer, 2022.

\bibitem{maple}
Muhammad~Uzair Khattak, Hanoona Rasheed, Muhammad Maaz, Salman Khan, and Fahad~Shahbaz Khan.
\newblock Maple: Multi-modal prompt learning.
\newblock In {\em Proceedings of the IEEE/CVF Conference on Computer Vision and Pattern Recognition}, pages 19113--19122, 2023.

\bibitem{promptsrc}
Muhammad~Uzair Khattak, Syed~Talal Wasim, Muzammal Naseer, Salman Khan, Ming-Hsuan Yang, and Fahad~Shahbaz Khan.
\newblock Self-regulating prompts: Foundational model adaptation without forgetting.
\newblock In {\em Proceedings of the IEEE/CVF International Conference on Computer Vision}, pages 15190--15200, 2023.

\bibitem{var_auto_encoder}
Diederik~P Kingma, Max Welling, et~al.
\newblock Auto-encoding variational bayes, 2013.

\bibitem{blip}
Junnan Li, Dongxu Li, Caiming Xiong, and Steven Hoi.
\newblock Blip: Bootstrapping language-image pre-training for unified vision-language understanding and generation.
\newblock In {\em International conference on machine learning}, pages 12888--12900. PMLR, 2022.

\bibitem{proda}
Yuning Lu, Jianzhuang Liu, Yonggang Zhang, Yajing Liu, and Xinmei Tian.
\newblock Prompt distribution learning.
\newblock In {\em Proceedings of the IEEE/CVF Conference on Computer Vision and Pattern Recognition}, pages 5206--5215, 2022.

\bibitem{star}
Zhihe Lu, Yongxin Yang, Xiatian Zhu, Cong Liu, Yi-Zhe Song, and Tao Xiang.
\newblock Stochastic classifiers for unsupervised domain adaptation.
\newblock In {\em Proceedings of the IEEE/CVF conference on computer vision and pattern recognition}, pages 9111--9120, 2020.

\bibitem{plant_disease}
Sharada~P Mohanty, David~P Hughes, and Marcel Salath{\'e}.
\newblock Using deep learning for image-based plant disease detection.
\newblock {\em Frontiers in plant science}, 7:215232, 2016.

\bibitem{bnns}
Radford~M Neal.
\newblock {\em Bayesian learning for neural networks}, volume 118.
\newblock Springer Science \& Business Media, 2012.

\bibitem{clip}
Alec Radford, Jong~Wook Kim, Chris Hallacy, Aditya Ramesh, Gabriel Goh, Sandhini Agarwal, Girish Sastry, Amanda Askell, Pamela Mishkin, Jack Clark, et~al.
\newblock Learning transferable visual models from natural language supervision.
\newblock In {\em International conference on machine learning}, pages 8748--8763. PMLR, 2021.

\bibitem{clap}
Julio Silva-Rodriguez, Sina Hajimiri, Ismail Ben~Ayed, and Jose Dolz.
\newblock A closer look at the few-shot adaptation of large vision-language models.
\newblock In {\em Proceedings of the IEEE/CVF Conference on Computer Vision and Pattern Recognition}, pages 23681--23690, 2024.

\bibitem{deepfake}
Jiahe Tian, Cai Yu, Xi~Wang, Peng Chen, Zihao Xiao, Jizhong Han, and Yesheng Chai.
\newblock Dynamic mixed-prototype model for incremental deepfake detection.
\newblock In {\em Proceedings of the 32nd ACM International Conference on Multimedia}, pages 8129--8138, 2024.

\bibitem{chestx}
Xiaosong Wang, Yifan Peng, Le~Lu, Zhiyong Lu, Mohammadhadi Bagheri, and Ronald~M Summers.
\newblock Chestx-ray8: Hospital-scale chest x-ray database and benchmarks on weakly-supervised classification and localization of common thorax diseases.
\newblock In {\em Proceedings of the IEEE conference on computer vision and pattern recognition}, pages 2097--2106, 2017.

\bibitem{semantic_guided}
Kangyu Xiao, Zilei Wang, and Junjie Li.
\newblock Semantic-guided robustness tuning for few-shot transfer across extreme domain shift.
\newblock In {\em European Conference on Computer Vision}, pages 303--320. Springer, 2024.

\bibitem{tcp}
Hantao Yao, Rui Zhang, and Changsheng Xu.
\newblock Tcp: Textual-based class-aware prompt tuning for visual-language model.
\newblock In {\em Proceedings of the IEEE/CVF Conference on Computer Vision and Pattern Recognition}, pages 23438--23448, 2024.

\bibitem{taskres}
Tao Yu, Zhihe Lu, Xin Jin, Zhibo Chen, and Xinchao Wang.
\newblock Task residual for tuning vision-language models.
\newblock In {\em Proceedings of the IEEE/CVF Conference on Computer Vision and Pattern Recognition}, pages 10899--10909, 2023.

\bibitem{re-id}
Tianyuan Yu, Da~Li, Yongxin Yang, Timothy~M Hospedales, and Tao Xiang.
\newblock Robust person re-identification by modelling feature uncertainty.
\newblock In {\em Proceedings of the IEEE/CVF international conference on computer vision}, pages 552--561, 2019.

\bibitem{tip_adapter}
Renrui Zhang, Rongyao Fang, Wei Zhang, Peng Gao, Kunchang Li, Jifeng Dai, Yu~Qiao, and Hongsheng Li.
\newblock Tip-adapter: Training-free clip-adapter for better vision-language modeling.
\newblock {\em arXiv preprint arXiv:2111.03930}, 2021.

\bibitem{stylematch}
Kaiyang Zhou, Chen~Change Loy, and Ziwei Liu.
\newblock Semi-supervised domain generalization with stochastic stylematch.
\newblock {\em International Journal of Computer Vision}, 131(9):2377--2387, 2023.

\bibitem{coop}
Kaiyang Zhou, Jingkang Yang, Chen~Change Loy, and Ziwei Liu.
\newblock Learning to prompt for vision-language models.
\newblock {\em International Journal of Computer Vision}, 130(9):2337--2348, 2022.

\bibitem{prompt_free}
Linhai Zhuo, Zheng Wang, Yuqian Fu, and Tianwen Qian.
\newblock Prompt as free lunch: Enhancing diversity in source-free cross-domain few-shot learning through semantic-guided prompting.
\newblock {\em arXiv preprint arXiv:2412.00767}, 2024.

\end{thebibliography}

% \input{ReproducibilityChecklist/LaTeX/ReproducibilityChecklist.tex}

\end{document}